# UTF-8 Plumbing: Byte-level Tokenizers Unavoidably Enable LLMs to Generate Ill-formed UTF-8

**Preston Firestone**   **Shubham Ugare**   **Gagandeep Singh**   **Sasa Misailovic**
University of Illinois Urbana-Champaign, USA

## Abstract

Subword tokenization segments input text according to a pre-defined vocabulary to feed it into a language model; the language model, in turn, generates a sequence made from this same vocabulary. The members of the vocabulary can be built of code points or bytes. Using code points means that all members of the vocabulary are valid UTF-8 characters. However, it also requires thousands of initial members to achieve acceptable coverage of inputs. Beginning with bytes, on the contrary, avoids out-of-vocabulary errors with only 256 initial members of the vocabulary, but the members of the vocabulary and sequences of them are not guaranteed to be valid UTF-8. Sequences that are not valid UTF-8 break code that assumes its input to be valid UTF-8. Applications of language models must account for the breakage thereby introduced. In this paper, we formalize tokenization using monoid theory and prove that tokenizers whose vocabularies contain tokens that are ill-formed UTF-8 can always produce sequences that are ill-formed UTF-8. We show formally that attempting to incrementally convert tokens back to a string and interpret the results as UTF-8 gives different results than converting the whole sequence of tokens at once. This formal result predicts real-world bugs: we evaluate mitigations for the problem identified and provide case studies of major foundation models and constrained generation systems.

## 1 Introduction

Large language models (LLMs) compute the probability of a given sequence of tokens drawn with replacement from a finite vocabulary. Tokenization is the process of cutting an input string into tokens such that each token is a member of a predefined vocabulary. Traditionally, tokenization used prior knowledge of the language, such as words, morphemes, or phonemes, to separate the text into useful elements (Grefenstette & Tapanainen, 1994; Palmer, 2000). Contemporary natural language processing has moved to using "subword" tokenization, where the vocabulary is derived automatically from a corpus of texts rather than programmed manually (Mielke et al., 2021).

The text the tokenizer works with is stored in computers as bytes. In order to interpret these bytes as characters meaningful to humans, we must define an encoding scheme mapping between a sequence of bytes and a sequence of characters. The predominant scheme is called UTF-8, regulated by Unicode Technical Committee (2025, §3.4, 3.9.3, 3.10). As of 2025, UTF-8 is used by 98.5% of all websites (W3Techs, 2025) and is required by the W3C for user agents (van Kesteren, 2024, §4.2).

Figure 1 shows two examples of strings as characters, bytes, and tokens, from different languages and models. Contemporary language models' tokenizers typically work with the bytes on the bottom row of Figure 1 rather than the characters on the top row. Abstracting away the bytes to focus on the characters is an instance of a "leaky" abstraction, a common issue in software engineering: the information that we tried to hide by an abstraction ends up being exposed (often as we need to handle errors) (see Spolsky, 2002; Kiczales et al., 1992; Kiczales, 1991).

| अ | ग |  | न | ि | म | ी | ळ | े |
|---|---|---|---|---|---|---|---|---|
| ⸨E0 A4⸩85⸩ | ⸨E0 A4⸩97⸩ | ⸨E0 A5 8D | E0 A4⸩A8⸩ | ⸨E0 A4 BF | E0 A4⸩AE⸩ | ⸨E0 A5 80⸩ | ⸨E0 A4⸩B3⸩ | ⸨E0 A5 87⸩ |

(a) Characters and bytes representing the opening of the *Rigveda* (Sanskrit: ऋग्वेद), "अग्निमीळे", transliterated *agnim īḻe* ("I praise Agni [fire]"), tokenized according to `cl100k_base`.

| Г | р | а | д |  | г | р | а | д | и | л | а |
|---|---|---|---|---|---|---|---|---|---|---|---|
| ⸨D0 93⸩ | ⸨D1 80 | D0 B0 | D0 B4⸩ | ⸨20 | D0 B3⸩ | ⸨D1 80 | D0 B0 | D0 B4⸩ | ⸨D0 B8 | D0 BB | D0 B0⸩ |

(b) The first two words of *The Building of Skadar* (Serbian Cyrillic: Зидање Скадра), "Град градила", transliterated *Grad gradila* ("The city was built"), tokenized according to Qwen3's vocabulary.

Figure 1: Examples of UTF-8 encoding. In each subfigure, the top row contains the individual Unicode characters that compose the string (see Unicode Technical Committee, 2025, §2.11), and the bottom row contains the UTF-8 bytes that encode each character in the string; each byte is represented by two hexadecimal numerals. The squiggles ⸨ indicate the boundaries between tokens.

Concretely, concatenating tokens made up of bytes does not always result in sequences that can be successfully interpreted as characters. In Figure 1a, the first token, ⸨E0 A4⸩, does not encode any character by itself; only when combined with the second token ⸨85⸩ can the resulting three bytes E0 A4 85 be interpreted as the character अ. Since language models generate sequences made up of the tokens in their vocabulary rather than generating bytes directly, programs that interact with sequences generated by a language model must handle byte sequences that cannot be interpreted as characters.

**Contributions.** In this paper, we make the following contributions:

⋆ Introduce a formal framework for tokenization based on monoid theory.
⋆ Prove, using our formalism, that vocabularies containing tokens that cannot be interpreted as characters can always produce sequences that cannot be interpreted as characters.
⋆ Show that interpreting bytes as UTF-8 characters is not a homomorphism and formally represent common mitigation strategies in the framework of monoids.
⋆ Classify popular language models according to the type of tokenizer they use.
⋆ Fix failures in constrained generation.

## 2 Background on monoids

To treat tokenizers abstractly, we introduce a formalism based on monoids, common in the literature on combinatorics on words (Sakarovitch, 2009; Lothaire, 1997) but to our knowledge new to natural language processing.

**(Free) monoids.** The basic component of our abstraction is the monoid.

**Definition 1** (Monoid). *A monoid is a triple composed of a set of elements, a binary operation on members of that set, and an identity element.* We refer to monoids by Greek capital letters ($\Sigma$), their members by subscripted lowercase Latin letters ($s_0, s_1 \in \Sigma$), the binary operation as a subscripted dot ($\cdot_\Sigma$), and the identity element as a subscripted epsilon ($\epsilon_\Sigma$). A **sequence** of members of a monoid is referred to by a lowercase Greek letter ($\sigma = s_0 \cdot_\Sigma s_1$). Subscripts and binary operations are left out when context makes the meaning clear.

Monoids' binary operation is associative but not commutative. For all members of the monoid,

$$(s_1 \cdot_\Sigma s_2) \cdot_\Sigma s_3 = s_1 \cdot_\Sigma (s_2 \cdot_\Sigma s_3). \tag{1}$$

The identity element is a neutral element for the binary operation, such that for any sequence $\sigma$

$$\sigma \cdot_\Sigma \epsilon_\Sigma = \epsilon_\Sigma \cdot_\Sigma \sigma = \sigma. \tag{2}$$

Set theoretical operations ($\subset, \subseteq, =, \ldots$) are defined between monoids that share a binary operation and an identity element.

An example monoid is a language, where the symbols that make up the language are the set of members $s_0 \ldots$, concatenation the binary operation $\cdot_\Sigma$, and the empty string the identity element $\epsilon_\Sigma$. In subsequent sections we will reason about all the sequences that can be formed from a given monoid, which make up the free monoid of a given monoid.

**Definition 2** (Free monoid). The **free monoid** generated by a monoid $\Sigma$, denoted $\Sigma^*$, is the monoid of all sequences of finite length made of elements from $\Sigma$. The free monoid is like the Kleene closure over the set of members of $\Sigma$, except the free monoid only contains finite sequences.

In Section 3, we will use the free monoid to describe the set of all sequences that can be generated with a given vocabulary, and in Section 4 we will use a free monoid to describe all byte sequences that can be interpreted as a string.

**Properties of monoids and their free monoids.** We will need to compare the contents of the free monoids of two different monoids that share a binary operation and identity element; this will allow us to make claims about what kinds of sequences are and are not in a given free monoid.

**Lemma 1.** *If some monoid $\Sigma$ contains a member not in some other free monoid $\Delta^*$, where $\Sigma$ and $\Delta^*$ share a binary operation and an identity element, then the free monoid $\Sigma^*$ of the first monoid $\Sigma$ also contains a members not in the other free monoid $\Delta^*$:*

$$\Sigma \nsubseteq \Delta^* \to \Sigma^* \nsubseteq \Delta^*.$$

*Proof.* Suppose for contradiction that there exist some $\Sigma$ and $\Delta$ such that $\Sigma \nsubseteq \Delta^*$ and $\Sigma^* \subseteq \Delta^*$; the existence of such a pair would be a counterexample to Lemma 1. For the set of members of $\Sigma$ to not be a subset of $\Delta^*$, there must exist some member of $\Sigma$ that is not in $\Delta^*$. That member of $\Sigma$ is also in $\Sigma^*$, but that implies that $\Sigma^* \nsubseteq \Delta^*$. This contradiction shows that no counterexample to Lemma 1 exists, thereby proving that the implication holds for all monoids $\Sigma$ and $\Delta$ that share a binary operation and an identity element. □

**Mapping between monoids.** In order to describe the process of tokenization in Section 3, we will need to map sequences back and forth between different monoids. To that end we introduce the abstraction of a stochastic map (Gastaldi et al., 2024).

**Definition 3** (Stochastic map). A **stochastic map** $\kappa$ from a monoid $\Sigma$ to a monoid $\Delta$ is a function from $\Sigma$ to the set of probability distributions on $\Delta$.

Other works analyzing tokenizers (e.g. Zouhar et al., 2023b; Geng et al., 2024; van Antwerpen & Neubeck, 2024) assume that the tokenizer is deterministic or have to cope with the ways it might not be. Following Gastaldi et al. (2024) we use stochastic maps to ensure that our results apply to nondeterministic tokenizers, which might result in more robust models (Geh et al., 2025; Chai et al., 2024a; Hofmann et al., 2022). Similarly, in our notation we do not explicitly depict the probability distribution over tokenizations that $\tau$ returns, because we do not examine the probability of individual tokenizations, but merely which tokenizations are or are not possible for a given input.

We introduce the property of homomorphism as a restriction on maps:

**Definition 4** (Homomorphism). A map $\kappa$ from a monoid $\Sigma$ to $\Delta$ is a **homomorphism** if and only if, for all $\sigma$, $\sigma'$ in $\Sigma$,

$$\kappa(\sigma \cdot_\Sigma \sigma') = \kappa(\sigma) \cdot_\Delta \kappa(\sigma'). \tag{3}$$

We show in Section 3 that tokenizing a string is not a homomorphism and in Section 4 that interpreting bytes as a string is also not a homomorphism. We then discuss real-world mitigations (Section 5) and bugs that arise because tokenizers fail to be homomorphisms (Section 6).

## 3 Tokenization described with monoids

Using the tools we introduced, we can describe a tokenizer as a pair of stochastic maps between two monoids and prove that tokenizing is not a homomorphism.

**Binding and cutting monoids.** We need to describe, in terms of monoids, the process of cutting that a tokenizer does to its input. The system we introduce in this subsection is similar but not identical to the system of Berglund & van der Merwe (2023). See Section 7 for a comparison of our system with theirs.

**Definition 5** (Cut monoid). The **cut monoid** of a given monoid is a monoid each of whose members has been prepended and postpended by a squiggle $\wr$. The cut monoid of a given monoid $\Sigma$ is indicated by a superscripted squiggle:

$$\Sigma^{\wr} = \{\wr \sigma \wr | \sigma \in \Sigma\}. \tag{4}$$

Where convenient to do so, we collapse adjacent squiggles $\wr$ for legibility.

For convenience, we use the notation $\Sigma^{\circ}$ to name a finite subset of some monoid $\Sigma$; we call the operator $\circ$ "bind". All three operators, cutting ($\wr$), binding ($\circ$), and freeing ($*$), can be applied in succession to a single monoid, which we depict by putting multiple superscripts on the capital Greek letter representing the monoid; the operations are applied in order from left to right.

We next illustrate our notation and define several monoids relevant in the rest of this paper. The monoid $\Sigma^*$ is all possible finite sequences of members of $\Sigma$. The monoid $\Sigma^{*\wr}$ is the set of all possible finite sequences of members of $\Sigma$, each sequence with a squiggle pre- and postpended to it—it represents the set of all possible tokens. $\Sigma^{*\wr\circ}$ is a finite subset of $\Sigma^{*\wr}$—it represents the vocabulary of some tokenizer. $\Sigma^{*\wr\circ*}$ contains all possible sequences of members of $\Sigma^{*\wr\circ}$—it represents all sequences that can be made using the vocabulary of a given tokenizer.

**What is a tokenizer?** Tokenization is cutting the input into tokens. We define it as a pair of mappings between two monoids $\Sigma^*$ and $\Sigma^{*\wr\circ*}$ that share a binary operation and identity element:

**Definition 6** (Tokenizer). We call a tokenizer over some free monoid $\Sigma^*$ and some vocabulary $\Sigma^{*\wr\circ}$ a pair of stochastic maps $\tau : \Sigma^* \to \Sigma^{*\wr\circ*}$ and $\kappa : \Sigma^{*\wr\circ*} \to \Sigma^*$.

The map $\tau$ cuts its argument such that each token of the cut sequence is a member of some the vocabulary $\Sigma^{*\wr\circ}$; $\kappa$ joins the cut sequence by removing all the cut operators $\wr$. $\kappa$ is deterministic and works the same way for any vocabulary made from the same initial monoid, whereas the behavior of $\tau$ depends on its vocabulary. Each tokenizer $(\tau, \kappa)$ is parameterized by the vocabulary $\Sigma^{*\wr\circ}$ it is defined over, but we omit this detail in the notation because we will only work with one vocabulary at a time. We will refer to passing some input through $\tau$ as **tokenizing** and some input through $\kappa$ as **detokenizing**.[1]

**Tokenizers and homomorphism.** In the paper, we will rely on the following property of tokenizers as we defined them in Definition 6:

**Theorem 1.** *Given a tokenizer $(\tau, \kappa)$, $\kappa$ is a homomorphism, but $\tau$ is not.*

*Proof. $\kappa$ is a homomorphism:* The result of concatenating two cut sequences and removing the cut marks is always identical to removing the cut marks and concatenating them: the cut symbols are removed and the resultant string is identical to its uncut version. We have shown that there are no counter examples to Equation 3 for $\kappa$.

*$\tau$ is not homomorphism:* Take some $s_0 \cdots s_n \in \Sigma^*$ and cut it into some $\wr s_0 \cdots s_{m-1} \wr s_m \cdot s_{m+1} \wr s_{m+2} \cdots s_n \wr$. It is impossible to separately cut $s_0 \cdots s_m$ and $s_{m+1} \cdots s_n$ such that there is not a cut after $s_m$ and before $s_{m+1}$; therefore the concatenation of any separate cutting of the two sequences $s_0 \cdots s_m$ and $s_{m+1} \cdots s_n$ must include the subsequence $s_m \wr s_{m+1}$. Cutting the entire sequence $s_0 \cdots s_n$ at once allows for sequences that do not contain $s_m \wr s_{m+1}$. We show that there exist counterexamples to Equation 3 for $\tau$. □

---

[1]Though the standard in the literature is to use "decode" for $\kappa$, we chose an alternative term "detokenizing" to make it clear that we are speaking of turning tokens into some sequence of members of the uncut monoid, a step we consider separately from the process of interpreting those bytes as a UTF-8 string.

Theorem 1 has already been proved by Geng et al. (2024), though they use a different approach from ours. We reprove the Theorem here to exercise our system and state it in our terms, and extensively use it in Section 4. Theorem 1 applies to any tokenizer that uses cutting, regardless of vocabulary, the order in which cuts are applied, and whether the tokenization is stochastic or not.

**Out-of-vocabulary errors.** Sometimes there does not exist any way to cut a sequence such that the resulting cut sequence is made up exclusively of members of the tokenizer's vocabulary. Often, however, at least part of the input can be cut into members of the vocabulary. There are three basic ways to deal with sections of the input that cannot be represented in the vocabulary: failing entirely, dropping the unacceptable section, and replacing the section with something else. In terms of our formalism:

1. **Failing entirely:** declaring $\tau$ to be undefined where its input cannot be mapped to $\Sigma^{*\wr\circ*}$;
2. **Dropping an unacceptable section:** replacing the sequences between squiggles $\wr$ that are not in $\Sigma^{*\wr\circ*}$ with $\wr\epsilon_\Sigma\wr$ (this is a special case of the replacement strategy where $\eta = \epsilon$);
3. **Replacing the section with other symbols:** mapping the objectionable bytes, according to some scheme, to one or more members of $\Sigma^{*\wr\circ}$ specially set aside for this purpose.[2]

A $\tau$ that uses any of these mitigations will not be a homomorphism, because none of the mitigations avoids the counterexample given in the proof of Theorem 1. UTF-8 implementations often elect the third option, replacing ill-formed byte sequences with a "replacement character", U+FFFD ([?]).

## 4 UTF-8 and monoids

**The UTF-8 encoding scheme.** In the UTF-8 encoding scheme (Unicode Technical Committee, 2025, §3.9.3), each character is paired with some byte sequence one to four bytes long; a single such byte sequence is called a UTF-8 **code unit**. Descriptions of the UTF-8 code units are found in Table 1; any sequence of bytes made up exclusively of concatenated, non-overlapping UTF-8 code units is a **well-formed** UTF-8 string. See Appendix B for more details about Unicode and UTF-8. The rest of this section will discuss UTF-8 using the monoid and tokenizer abstractions we introduced earlier. We begin by phrasing the encoding of characters as bytes and the interpretation of bytes as characters in terms of monoids.

| First byte | Second byte | Third byte | Fourth byte |
|---|---|---|---|
| 00..7F | | | |
| C2..DF | 80..BF | | |
| E0 | A0..BF | 80..BF | |
| E1..EC | 80..BF | 80..BF | |
| ED | 80..9F | 80..BF | |
| EE..EF | 80..BF | 80..BF | |
| F0 | 90..BF | 80..BF | 80..BF |
| F1..F3 | 80..BF | 80..BF | 80..BF |
| F4 | 80..8F | 80..BF | 80..BF |

Table 1: Well-formed UTF-8 byte sequences (Unicode Technical Committee, 2025, §3.9.3). A range such as 80..BF should be read as an inclusive range from 80 through BF.

**Definition 7** (UTF-8)**.** We refer to the monoid of UTF-8 code units as Y, which is the $B^{*\wr\circ}$ specified in Table 1. The sequences of $B^*$ that are not in $Y^*$ are ill-formed UTF-8 sequences; the members of $B^*$ that are in $Y^*$ are well-formed.

Not all byte sequences are well-formed UTF-8; see Appendix B for examples. The existence of byte sequences that are ill-formed in UTF-8 motivates Theorem 2, which follows immediately from Lemma 1 and Definition 7.

**Theorem 2.** *Any vocabulary $B^{*\wr\circ}$ that contains a byte sequence $\beta$ that is ill-formed in UTF-8 ($\beta \notin Y^*$) will be able to generate ill-formed sequences:*

$$B^{*\wr\circ} \nsubseteq Y^* \to B^{*\wr\circ*} \nsubseteq Y^*.$$

---

[2]See Unicode Technical Committee (2025, §3.9.6) and van Kesteren (2024, §4.1) for the standard algorithm for this replacement.

Due to Theorem 2, any language model's vocabulary that contains byte tokens ill-formed in UTF-8 can generate token sequences that are ill-formed in UTF-8.

**Enforcing UTF-8 breaks homomorphism in tokenizers.** In Section 6 we will show impacts of assuming that the output of a language model is well-formed UTF-8. To prove that the problems we discuss are inevitable, we introduce the following theorem.

**Theorem 3.** *Interpreting a byte sequence as UTF-8 is not a homomorphism, but serializing a sequence of well-formed UTF-8 byte sequences is a homomorphism.*

*Proof.* Mapping between bytes and well-formed UTF-8 sequences is like tokenization, as in Definition 6. Mapping from well-formed UTF-8 sequences to a sequence of bytes means simply concatenating those bytes (Unicode Technical Committee, 2025, §3.10). Joining the UTF-8 code units into a byte string is naturally like a detokenizer $\kappa$ from $Y^*$ (i.e. $B^{*\ell \circ *}$) to $B^*$. Mapping from a sequence of bytes to a well-formed UTF-8 sequence requires splitting that byte sequence into consecutive, adjacent UTF-8 code units. This is like a $\tau$ from $B^*$ to $Y^*$ (i.e. to $B^{*\ell \circ *}$). Theorem 3 follows directly by application of Theorem 1. □

Theorem 3 applies regardless of which of the strategies of Section 3 is used. In the next section, we will assume that decoding applies the third strategy discussed in Section 3, and we will refer to the member of Y set aside for ill-formed byte sequences as $\eta$. The number of $\eta$s used to replace ill-formed bytes in the sequence and their identity is variable. In Section 6 we will see examples where more than one distinct replacement character exists. See Unicode Technical Committee (2025, §3.9.6) for the standard algorithm used for replacement when decoding UTF-8. The process assumes a single $\eta$, the character U+FFFD "", known as a *replacement character*. The number of $\eta$s introduced to suture any ill-formed sections of the byte sequence will not be relevant to the results of the next section, only that such a unique character exists.

## 5 Incrementally detokenizing ill-formed UTF-8 sequences

It is common for language model applications to stream or otherwise incrementally generate tokens, decoding and displaying the decoded tokens opportunistically. Interactive systems, for example, can display the intermediate steps of the generation to the user; hosted language models on a remote server might send tokens to the client as soon as they become available, rather than wait for generation to complete before sending anything downstream.

As we have shown in Section 4, certain combinations of tokens can be ill-formed UTF-8, causing a process that attempts to decode them to apply one of the coping strategies of Subsection 3. The decoding processes used in deployed tokenizers use the third strategy described at the end of Section 3: "replacing the [ill-formed] section with other symbols", in particular they use the symbol U+FFFD  ("replacement character"); our formalism calls it $\eta$.

If the user attempts to interpret the byte tokens as UTF-8 as the tokens come in, then concatenate, the result will be different from what the user would have gotten had they waited for all the tokens before interpreting them all together. This problem was detected in serving engines for language models, which host language models and serve them through an API locally or over the network: these engines failed to generate correct text, because they incorrectly assumed that the conversion from bytes to well-formed UTF-8 was a homomorphism. This problem and a common mitigation strategy were introduced to Hugging Face TGI by a user's issue (Hugging Face, 2025; 0x1997, 2023) and percolated thence to vLLM (Kwon et al., 2023; Yard1, 2023), OpenLLM (bentoml, 2025; jeffwang0516, 2023), and SGLang (Zheng et al., 2024; hnyls2002, 2024), all of which faced similar issues for the same reason. See Appendix C for a Python code from the serving engines.

**Algorithm 1** Incrementally generate using byte-level tokenizer.

**Require:**
- $\tau : B^* \to Y^*$. Tokenize a sequence of bytes to a sequence of UTF-8 code units, replacing ill-formed bytes with $\eta$.
- $\kappa : B^{*\ell\circ*} \to B^*$. Detokenize from a sequence of byte tokens in some vocabulary $B^{*\ell\circ}$ to a sequence of bytes.
- $LM : B^{*\ell\circ*} \to B^{*\ell\circ}$. A language model that takes in a sequence of tokens and returns a new token.
- $i$ : stateful integer initialized to 0.
- $j$ : stateful integer initialized to 0.

1: **function** ADVANCETOKEN( $\beta : B^{*\ell\circ*}$)
2:     $v : Y^* \leftarrow \tau(\kappa(\beta[i:j]))$
3:     $v' : Y^* \leftarrow \tau(\kappa(\beta[i:\text{LENGTH}(\beta)]))$
4:     **if** $\text{LENGTH}(v') > \text{LENGTH}(v) \land$
         $v'[\text{LENGTH}(v') - 1] \neq \eta$ **then**
5:         $i \leftarrow j$
6:         $j \leftarrow \text{LENGTH}(\beta)$
7:         **return** $v'[\text{LENGTH}(v) : \text{LENGTH}(v')]$
8:     **else**
9:         **return** $\epsilon_Y$

1: **function** GENERATE
2:     $\beta : B^{*\ell\circ*} \leftarrow \epsilon_B$
3:     $v : Y^* \leftarrow \epsilon_Y$
4:     **loop**
5:         $\beta \leftarrow \beta \cdot_B LM(\beta)$
6:         $v \leftarrow v \cdot_Y \text{ADVANCETOKEN}(v, \beta)$
7:         EMIT($\beta, v$)

Algorithm 1 formalizes this mitigation strategy using the framework of monoids. The function ADVANCETOKEN opportunistically interprets byte tokens as UTF-8 sequences when the end of the sequence's final token is aligned with the edge between two code points. Each time it is called, ADVANCETOKEN receives all the tokens generated so far, with the new token at the end of the sequence. If there is a new well-formed code unit at the end of the sequence of tokens, ADVANCETOKEN updates $i$ and $j$ to point to the last two positions in the token sequence where the token and code unit boundary lined up and returns the new successfully-decoded code units. The algorithm ignores the tokens before the $i^{\text{th}}$ token, because appending bytes to well-formed UTF-8 cannot change the previous code units: well-formed UTF-8 code units do not overlap.

The outer function, GENERATE, shows how to exercise ADVANCETOKEN to emit the incrementally produced text as well-formed code units are completed. Both the byte tokens and text generated so far are stored in the local variables $\beta$ and $v$, which enables the generating process to output text as it is generated without losing information about the exact bytes the model generated. Note the binary operations on lines 5 and 6 of the function GENERATE in Algorithm 1: the first concatenates bytes and the second UTF-8 code units. The purpose of Algorithm 1 is to enable the simultaneous application of these operations without destructively decoding the byte tokens such that further tokens cannot be successfully appended to the sequence.

Algorithm 1 cannot make concatenating UTF-8 tokens a homomorphism (consistent with Theorems 1 and 3). If one attempted to concatenate two sequences in $Y^*$ that came out of the function GENERATE, one would always end up with a sequence in $Y^*$. But if the sequences in $B^{*\ell\circ*}$ that Algorithm 1 detokenizes and decodes as UTF-8 are not well-formed UTF-8, then the outputs of Algorithm 1 will remove or mangle the information the ill-formed bytes contained. For example, the text of Figure 1a contains twelve tokens. Decoding the first five with Algorithm 1 produces the UTF-8 code units "अग"; decoding the remaining seven produces "?मीळे": concatenating produces "अग?मीळे", not the expected "अग्नमीळे". Dropping the ill-formed bytes, strategy 2 from Section 3, will not restore homomorphism: "अग" · "मीळे" is "अगमीळे", not "अग्नमीळे". Algorithm 1 is not always necessary: none of the Cyrillic characters in Figure 1b are split across more than one token, while several of the Devanagari characters Figure 1a are. Detokenizing and decoding this Cyrillic sequence is a homomorphism as a special case, because all of its tokens are well-formed; correctly handling the Devanagari sequence, on the other hand, requires special handling.

The information lost when decoding an ill-formed byte sequence (whether failing, dropping, or replacing) is always gone. Concatenating the decoded tokens will therefore produce different results from decoding the concatenated tokens wherever the split between tokens is not aligned with the split between code units. Algorithm 1 decodes the intermediate results of generation while correctly appending new tokens to those generated so far. The local variable $v$ in the function GENERATE always contains the code units generated so far; if the final code unit is the filler character $\eta$ (?), then $v$ does not contain the final code unit. Because it appends in the tokens' monoid $B^{*\lozenge}$ as well as in the code units' monoid Y (see lines 5 and 6 of the function GENERATE), Algorithm 1 is a partial fix for certain cases, but as we have seen, it has inherent limitations.

## 6  Sealing the leaks in UTF-8 decoding

Deployed tokenizers usually concatenate byte tokens and interpret them as if they were UTF-8; when the process reaches an ill-formed subsequence, it has the choices outlined in Section 3: failure, dropping, and replacement. Theorem 3 shows that if a vocabulary contains byte tokens that are ill-formed UTF-8, it will be able to produce sequences that are ill-formed UTF-8. We next classify the tokenizers of common foundation models and discuss coping with the impacts of Theorem 3 in constrained generation.

**Tokenization strategies of foundation models.** In Table 2, we classify several popular language models according to the style of tokenizer they use. We find two broad categories: byte-level, and byte-fallback. Byte-level tokenizers impose no constraints on the formation of tokens, which may or may not be valid in UTF-8. Since all non-ASCII characters are represented by more than one byte in UTF-8, any script that uses a character other than the 128 in ASCII could have its characters split across multiple tokens by a byte-level tokenizer.

Byte-fallback tokenizers require that all tokens in the vocabulary be valid in UTF-8, with the exception of 256 (or 243, excluding the 13 bytes that never appear in any UTF-8 code unit) tokens, one for each byte. These extra tokens are used to represent parts of the input that cannot otherwise be represented by tokens in the tokenizer's vocabulary. Figure 2 gives an example of such a tokenization, where each of the tokens contains a valid UTF-8 string, except for the three single-byte tokens used to encode the rare character ꙮ ("multiocular o", used in a single manuscript from the 15th century). Theorem 2 applies to both strategies equally, because the single-byte tokens are already ill-formed and so a sufficient condition for the theorem.

| м | ного | ꙮ | чит | ї | й |
|---|---|---|---|---|---|
| ⟨D0 BC⟩ | ⟨D0 BD D0 BE D0 B3 D0 BE⟩ | ⟨EA⟩⟨99⟩⟨AE⟩ | ⟨D1 87 D0 B8 D1 82⟩ | ⟨D1 97⟩ | ⟨D0 B9⟩ |

Figure 2: Characters and bytes representing the Old Church Slavonic word многоꙮчитїй, transliterated *mnogoočitii* ("many-eyed"), tokenized according to the vocabulary of Gemma-3.

Eight out of ten of the model families classified in Table 2 retain the same style of tokenization throughout their generations. Llama changes from SentencePiece (Kudo & Richardson, 2018) to a vocabulary derived from OpenAI's `tiktoken` (OpenAI, 2025) for its third generation (Grattafiori et al., 2024). This change of dependencies resulted in a switch from byte-fallback (since SentencePiece enforces UTF-8 validity on its tokens by default, the only mitigation it offers for covering out-of-vocabulary inputs is byte-fallback) to byte-level. Phi-3, unlike the other Phi-series models, used Llama 2's vocabulary; Phi-{1, 1.5} use vocabu-laries from CodeGen; Phi-2's documentation does not describe the tokenization in detail, but manual inspection of the publicly-available vocabulary reveals it to be byte-level; and Phi-4 uses one of OpenAI's vocabularies from `tiktoken` (OpenAI, 2025), all of which are byte-level. As we discuss in Section 7, ablations are seldom performed on tokenizers, and none of the models in Table 2 were tested with more than one tokenizer. We presume that the reason for skipping ablations is the excessive cost of training multiple models of the same scale and architecture but with distinct tokenizations (and embeddings).

| Model | Tokenizer Type |
|---|---|
| OpenAI since GPT-2 (Radford et al., 2019; Brown et al., 2020; OpenAI et al., 2024) | Byte-level |
| Qwen, Qwen2.5, Qwen3 (Bai et al., 2023; Yang et al., 2025b;a) | Byte-level |
| Llama 1, 2 (Touvron et al., 2023a;b) | Byte-fallback |
| Llama 3 (Grattafiori et al., 2024) | Byte-level |
| Mistral, Mixtral (Jiang et al., 2023; 2024) | Byte-fallback |
| Gemma 1, 2, 3 (Mesnard et al., 2024; Gemma Team et al., 2024; 2025) | Byte-fallback |
| OLMo, OLMo 2 (Groeneveld et al., 2024; Team OLMo et al., 2025) | Byte-level |
| Phi-1, 1.5, 2, 4 (Gunasekar et al., 2023; Li et al., 2023; Javaheripi & Bubeck; Microsoft et al., 2025) | Byte-level |
| Phi-3 (Abdin et al., 2024) | Byte-fallback |
| Cohere R and R+ (Cohere, 2024) | Byte-level |
| Stable LM 1, 2 (Stability-AI; Bellagente et al., 2024) | Byte-level |
| CodeGen, CodeGen2 (Nijkamp et al., 2023b;a) | Byte-level |

Table 2: Foundation models and the tokenizers they use.

**Existing constrained generation systems and non-homomorphic tokenizers.** Constrained generation techniques are used to restrict language model outputs to adhere to specified rules. We examined various grammar-constrained generation systems and tested them to discover their behavior during partial generation decoding (Scholak et al., 2021; Poesia et al., 2022; Willard & Louf, 2023; Ugare et al., 2025b;a; Banerjee et al., 2025; Loula et al., 2025; Suresh et al., 2025). Among popular grammar-constrained generation tools, we found that Synchromesh (Poesia et al., 2022) and SynCode (Ugare et al., 2025b) encountered issues when grammars included non-ASCII characters such as emojis or mathematical symbols such as '$\forall$'. Both tools use character-based parsers rather than byte-based ones, which created this vulnerability.

The way to fix this problem is to constrain generation at the level of bytes rather than characters. Attempting to constrain at the level of characters fails where the bytes of the token are ill-formed UTF-8. For example, the character $\forall$, used in Lean (Moura & Ullrich, 2021), might be tokenized ⟨E2 88⟩ ⟨80⟩. If the model has generated the first token of the character, then the constrained generation algorithm must permit the second token as a continuation, even though neither of the tokens is well-formed. Constraining at the level of bytes rather than characters by using bytes as the input alphabet for lexers, DFAs, and so forth repairs this problem.

For the v0.3.0 release of SynCode, we replaced the previous character-level finite state machines with byte-level ones. To evaluate SynCode's ability to handle non-ASCII Unicode characters, we conducted an experiment using an emoji generation task. We selected a subset of the TweetEval emoji dataset Barbieri et al. (2020), filtering for three common emoji classes. The task required the model to generate exactly one emoji character in response to a given tweet, adhering to a constrained grammar specification (see listing 4 in Appendix E). We evaluated this task across 100 examples from the TweetEval test set on two versions of SynCode: v0.2.0 which used character-level constraints, and the current version which implements byte-level constraints. Performance improvements are reported in Table 3: v0.2.0 crashed on all examples, whereas v0.3.0 successfully generated emojis for all examples.

| Metric | SynCode v0.2.0 | SynCode v0.3.0 |
|---|---|---|
| Accuracy | 0% | 62% |
| Crash Rate | 100% | 0% |

Table 3: Performance comparison between SynCode versions on emoji generation task

## 7 Related work

We direct the reader to Appendix A for more related work.

**Formal approaches to tokenization.** Following Sakarovitch (2009) and Lothaire (1997), we treat languages as monoids. Monoids have found use in the analysis of formal languages (Yang & Wu, 2023) but hardly at all in the study of natural languages. The work of Joachim Lambek on type grammar has led him to study natural language in terms of ordered monoids (Lambek, 1997; 2007), but his work came to our attention too late to significantly impact our work here.

**UTF-8 challenges in tokenization.** Rahman et al. (2024, §II.D) discuss the challenges of UTF-8, both because it encodes characters as sequences of varying length, and many characters are encoded by more than one byte. The latter challenge cannot be avoided unless one wants to limit one's character space to 256 unique characters, but the former issue is unique to UTF-8 and distinguishes it from other encoding schemes, UTF-16 and UTF-32, which encode all code points in two and four bytes respectively.

**Ablations.** To our knowledge, ablations on the relative efficacy of byte-fallback and byte-level tokenization have not been performed. Dagan et al. (2024) ablate on datasets, pre-tokenization schemes, and vocabulary size; Ali et al. (2024) ablate on datasets, tokenization algorithms (BPE v. Unigram) and implementations (Hugging Face v. SentencePiece). None directly address the tradeoffs between byte-level and byte-fallback tokenization.

**Glitch tokens.** "Glitch tokens" are under-trained tokens in the model's vocabulary that can cause erratic behavior (Land & Bartolo, 2024; Geiping et al., 2024). They are distinct from the ill-formed UTF-8 sequences we discuss. Land & Bartolo (2024) discuss what they call "partial UTF-8" tokens, defined as "tokens representing byte sequences that cannot be converted to Unicode characters [decoded, in our terms] as they contain only part of the full UTF-8 encoding for a character". This is a special case of our ill-formed tokens: ill-formed byte sequences need not contain any part of a UTF-8 encoding form. Ill-formed tokens are explicitly excluded from Land & Bartolo (2024)'s experiments because they "are not suitable for building verification prompts", presumably because these prompts must be well-formed UTF-8 to work with existing interfaces (e.g. Hugging Face's).

The byte-fallback tokens (see Section 6) that encode the bytes of ASCII characters are a source of glitchy tokens (see Geiping et al., 2024, Figure 23, in the Appendix). Geiping et al. (2024) exclude byte-fallback tokens from the in the main body of the paper that reports the glitchiest tokens in the Llama 2 vocabulary, because the tokens that represent the bytes of the ASCII characters cannot be the result of ordinary tokenization.

**Improbable bigrams.** Jang et al. (2024) examine pairs of tokens made up of tokens that are not well-formed UTF-8 but that, when concatenated, make a well-formed byte sequence. These tokens enable make attacks similar to those of Land & Bartolo (2024); Geiping et al. (2024), with the difference that Jang et al. (2024) examine pairs of tokens, whereas Land & Bartolo (2024); Geiping et al. (2024) examine individual tokens. These bigrams are well-formed and distinct from our examination of ill-formed sequences.

**Superscripted squiggles.** We borrow our notation of squiggles $\wr$ for the division between tokens from Berglund & van der Merwe (2023), though we slightly redefine the notation of a superscripted squiggle: our cutting operation $\wr$ describes the set of all possible tokens made from a given monoid, whereas the $\Sigma^\wr$ of Berglund & van der Merwe (2023) is "the set of all tokenizations constructed from strings in $\Sigma^*$". Their $\Sigma^\wr$ is equivalent to our $\Sigma^{*\wr*}$. Also, we add binding ($\circ$) and introduce the application of multiple superscripts to a single base monoid.

## 8 Conclusions and future work

The paper shows that UTF-8, tokens, and strings are leaky abstractions when generating text using language models. Rather than burying the issue, practitioners should accustom themselves to not expecting the inputs or outputs of their LLM system to be well-formed UTF-8. Implementers of systems and applications for language models should test their implementations on non-ASCII characters and ensure that they behave properly when an input or generated sequence is ill-formed UTF-8. Authors should be more precise when specifying the tokenization scheme their model uses when describing its architecture. We leave for future work surveying more foundation models, testing language model applications, expanding Algorithm 1 for concatenation in both directions rather than just appending (though we note that it behaves correctly when appending an arbitrarily large number of new tokens), and experimenting with multiple constrained generation algorithms.

All humans have the right to interact with computers in their own language and to be able to use computers to generate and manipulate text, with equal regard to all languages. This paper works toward that goal by clarifying some common issues present with large language models. It has addressed a leak in one of the abstractions used to work with language, but the general problem of how to process all natural languages uniformly remains open.

## Acknowledgments

We thank the anonymous reviewers for their comments. This research was supported in part by NSF Grants No. CCF-1846354, CCF-2313028, CCF-2238079, CCF-2316233, CNS-2148583, and an Amazon Research Award.

## Ethics statement

Davis & Suignard (2014) describes security issues that face systems that interact with Unicode as closely as language model applications and infrastructure must. They describe non-visual exploits such as buffer overflows during encoding or decoding, text comparison, ill-formed input bytes, including several exploits that are particular to UTF-8. Davis & Suignard (2014)'s visual exploits are based on visual spoofs, visually mistakable strings: these are two or more different sequences of code points that appear the same to the user (see Unicode Technical Committee (2025, §3.11) and Davis et al. (2024)). These would be tokenized differently by all tokenizers, because they are not the same byte or code point sequence and so cannot be represented by the same tokens.

Visual spoofs can be used for attacks similar to those recently studied by Geh et al. (2025), where models generated radically different responses to varying tokenizations of the same prompt; in some cases Geh et al. (2025) broke safety and alignment restrictions trained into models. A visual spoof would circumvent a mitigation Geh et al. (2025) suggest for their attack: providers of language models as a service could prohibit users from tokenizing their own texts and require them to submit well-formed UTF-8 in order to avoid adversarial tokenizations. Visual spoofs could produce the same effect of adversarial tokenization without the user needing to have direct control over the tokenization process.

When discussing how to decode UTF-8 sequences, the Unicode standard offers the following foreboding words: "[s]ilently ignoring ill-formed sequences is strongly discouraged because joining text from before and after the ill-formed sequence can cause the resulting text to take a new meaning. This result would be especially dangerous in the context of textual formats that carry embedded program code" (Unicode Technical Committee, 2025, C10). This warning is particularly apt for processes executing code that a language model generated.

## A  Extended related work

**Tokenization notation and terminology.** The process we call "cutting" is a standard step in tokenization but is represented variously in the literature. Sennrich et al. (2016) use (|) or a space; Gastaldi et al. (2024) use (ı); Schmidt et al. (2024) and Bostrom & Durrett (2020) use a space; Koo et al. (2024) use (|); Kudo (2018) uses (/); Kudo & Richardson (2018) wrap tokens with square brackets; Cognetta & Okazaki (2024) use ␣; and Geng et al. (2024) use color to distinguish tokens. We follow Berglund & van der Merwe (2023) in using (⌇), though we extend their notation by requiring that cut sequences begin and end with (⌇).

**Morphisms in tokenization.** We use the term homomorphism the way Geng et al. (2024) use it. For Sakarovitch (2009), a homomorphism is a bijective morphism. Gastaldi et al. (2024) call *multiplicative* what we refer to as a homomorphism; with the following additional constraint that it map non-empty sequences to non-empty sequences, it also has a *trivial kernel*: $\delta \neq \epsilon_\Delta \rightarrow \kappa(\delta) \neq \epsilon_\Sigma$. Lothaire (1997) and Sakarovitch (2009) call a *morphism* what we call a homomorphism with the additional requirement that it map the identity member to the identity member: $\kappa(\epsilon_\Delta) = \epsilon_\Sigma$. We include neither of these additional restrictions because they are not necessary for our work here.

**Subword tokenization methods.** A cutting tokenizer like ours is a generalization of a subword tokenizer (Sennrich et al., 2016; Kudo, 2018; Kudo & Richardson, 2018). The performance of subword tokenizers is contested. The reader is directed *inter alia* to Gallé (2019); Zouhar et al. (2023a) for a favorable impression and to Bostrom & Durrett (2020); Schmidt et al. (2024); Chai et al. (2024a) for a negative one. Practically representing the negative camp, tokenizer-free models work directly on input characters (e.g. Tay et al., 2022; Clark et al., 2022), bytes (e.g. Xue et al., 2022; Pagnoni et al., 2024), or on images of input text (e.g. Salesky et al., 2021; Rust et al., 2023; Chai et al., 2024b).

**Byte-level and character-level approaches.** Byte-pair encoding has been studied extensively: Bostrom & Durrett (2020) argue that Unigram is superior to BPE. Gallé (2019) argue that BPE performs highly because it compresses the input, but Schmidt et al. (2024) perform experiments that suggest that compression is not necessary or sufficient for performance in a tokenizer. Zouhar et al. (2023a) propose an information theoretic standard, Rényi entropy, for why certain tokenizers perform better than others; Cognetta et al. (2024) supply counter examples to the argument of Zouhar et al. (2023a). Libovický et al. (2022) examine the efficacy of tokenizationless character-level machine translation compared to character-level BPE and find that the former performs at best only as well as the latter.

**Inter-language performance comparisons.** Limisiewicz et al. (2024); Hofmann et al. (2022) examine adapting subword tokenization to respect morphological boundaries in language in order to improve performance in non-European languages. Petrov et al. (2023); Ahia et al. (2023) show that byte-level and character-level tokenization introduce severe discrepancies among languages, due in part to the varying lengths of byte sequences used to represent text in various languages.

## B  Details about UTF-8

Table 4: UTF-8 bit distribution, showing how to convert code points, represented as binary numbers, into the bytes of UTF-8. (Unicode Technical Committee, 2025, Table 3-6).

| Code Point | First Byte | Second Byte | Third Byte | Fourth Byte |
| --- | --- | --- | --- | --- |
| 00000000 0xxxxxxx | 0xxxxxxx | | | |
| 00000yyy yyxxxxxx | 110yyyyy | 10xxxxxx | | |
| zzzzyyyy yyxxxxxx | 1110zzzz | 10yyyyyy | 10xxxxxx | |
| 000uuuuu zzzzyyyy yyxxxxxx | 11110uuu | 10uuzzzz | 10yyyyyy | 10xxxxxx |

In the main body of the paper we ignore the concept of code points entirely and exclusively treat UTF-8 code units. However, in Appendix D we discuss a hack to store byte-level tokens in well-formed UTF-8 files; we will need to discuss code points as such there. Each character in Unicode (Unicode Technical Committee, 2025) is mapped to a number in the range 0 to $10FFFF_{16}$, called a code point. To turn a code point into its UTF-8 code unit, convert the code point to bytes as shown in Table 4. A process that interprets UTF-8 bytes as characters, a process we waved away in the body of the paper, must disassemble the bytes it receives on its input to recover the code point of the character it must display.

Several properties of UTF-8 referenced in the main text are immediately obvious by visual inspection of Tables 4 and 1. For example, any sequence made up exclusively of bytes starting 10, that is, in the range 80..BF, can never be well-formed UTF-8. Similarly, no sequence made up exclusively of bytes beginning with 110, 1110, or 11110 can ever contain any well-formed encoded forms. Also, the bytes C0-C1 and F5-FF never appear in UTF-8 at all, so they can also be used to disrupt well-formed encoding forms.

Note also that not all characters are encoded by the same number of bytes. In tokenizers based on merging pairs of tokens, the number of bytes that make up a character could influence the likelihood that that character is represented by more than one token. We leave verifying or disproving this conjecture to future work.

## C  Python implementation of Algorithm 1

**Listing 1** Implementation of Algorithm 1 from the serving engines discussed in Section 5. In the real listing, the variable `replacement_char` is assigned the value �. Note that the variables we call $i$ and $j$ are named `prefix_offset` and `read_offset` respectively. The fast version of the Hugging Face `tokenizers` library uses a Rust translation of this logic (Huggingface, 2025).

```python
def decode_token(
        self,
        all_input_ids: List[int],
        prefix_offset: int = 0,
        read_offset: int = 0,
        skip_special_tokens: bool = False,
    ) -> Tuple[str, int, int]:
        prefix_text = self.tokenizer.decode(
            all_input_ids[prefix_offset:read_offset],
            skip_special_tokens=skip_special_tokens,
        )
        new_text = self.tokenizer.decode(
            all_input_ids[prefix_offset:],
            skip_special_tokens=skip_special_tokens
        )

        if len(new_text) > len(prefix_text) and not new_text.endswith(replacement_char):
            new_text = new_text[len(prefix_text) :]
            return new_text, read_offset, len(all_input_ids)
        else:
            return "", prefix_offset, read_offset
```

Table 5 gives an example of incrementally decoding a sequence of byte tokens. The first column of the execution trace contains the byte tokens to be detokenized, read in order from top to bottom. The second column shows the characters emitted as they are completed. The third and fourth columns show the values of $i$ and $j$ as they are advanced through the sequence of byte tokens.

Table 5: Trace of executing Algorithm 1 on the tokens from Example 1a.

| Byte token | Text emitted | i | j |
|---|---|---|---|
| ⟨E0 A4⟩ | | 0 | 0 |
| ⟨85⟩ | अ | 0 | 2 |
| ⟨E0 A4⟩ | | 0 | 2 |
| ⟨97⟩ | ग | 2 | 4 |
| ⟨E0 A5 8D E0 A4⟩ | | 2 | 4 |
| ⟨A8⟩ | न | 4 | 6 |
| ⟨E0 A4 BF E0 A4⟩ | | 4 | 6 |
| ⟨AE⟩ | मि | 6 | 8 |
| ⟨E0 A5 80⟩ | ी | 8 | 9 |
| ⟨E0 A4⟩ | | 8 | 9 |
| ⟨B3⟩ | ळ | 9 | 11 |
| ⟨E0 A5 87⟩ | े | 11 | 12 |

## D  Mapping bytes to Unicode and back

| Г | р | а | д | | г | р | а | д | и | л | а |
|---|---|---|---|---|---|---|---|---|---|---|---|
| ⟨D093⟩ | ⟨D180 | D0B0 | D0B4⟩ | ⟨20 | D0B3⟩ | ⟨D180 | D0B0 | D0B4⟩ | ⟨D0B8 | D0BB | D0B0⟩ |
| ⟨Đĵ⟩ | ⟨ÑĢ | Đ° | Đ´⟩ | ⟨Ġ | Đ³⟩ | ⟨ÑĢ | Đ° | Đ´⟩ | ⟨Đ¸ | Đ» | Đ°⟩ |

Table 6: The first two words of *The Building of Skadar* (Serbian: Зидање Скадра), ``Град градила'', transliterated *"Grad gradila"* ("The city was built"). The top row is the sequence of characters making up these words, the middle row is the sequence of bytes that are the UTF-8 encoding of this sequence of code points, and the bottom row is the bytes of the middle row mapped to characters according to the mapping defined in Listing 2. The squiggles ⟨ in the bottom two rows indicate the tokenization according to Qwen3's vocabulary (Yang et al., 2025a).

**Listing 2** Code from GPT-2 (Radford et al., 2019) that defines an injective but not surjective mapping between the set of natural numbers from 0 to 255 and the set of natural numbers. The numbers $21_{16}, \ldots 7E_{16}, A1_{16}, \ldots AC_{16}, AE_{16}, \ldots, FF_{16}$ are mapped to themselves. The remaining numbers are mapped in order to $100_{16}$ through $143_{16}$. The returned dictionary maps from `int`s representing byte values to `str`s of one character representing code points.

```python
def bytes_to_unicode():
    bs = (
        list(range(ord("!"), ord("~") + 1))
        + list(range(ord("¡"), ord("¬") + 1))
        + list(range(ord("®"), ord("ÿ") + 1))
    )
    cs = bs[:]
    n = 0
    for b in range(2**8):
        if b not in bs:
            bs.append(b)
            cs.append(2**8 + n)
            n += 1
    cs = [chr(n) for n in cs]
    return dict(zip(bs, cs))
```

Hugging Face distributes tokenizer vocabularies as UTF-8-encoded files. [3] Since programmers tend to work with UTF-8-encoded strings rather than directly with byte sequences, it is convenient to be able to represent arbitrary byte sequences, such as those produced by arbitrarily clumping bytes, as UTF-8 sequences. [4] GPT-2 (Radford et al., 2019), because it had a byte-level vocabulary stored in UTF-8 files, introduced a mapping between each byte and a code point (see B) and storing the code points as UTF-8 byte sequences. The tokenizer's implementation must convert back and forth between the code points and the bytes they represent to interact with byte-level inputs.

---

[3] In HTTP-speak: `content-type: text/plain; charset=UTF-8`

Listing 2 gives a Python procedure for mapping from bytes to (printable) code points. Most of the code points between $00_{16}$ and $FF_{16}$ are printable characters, so though the result will be an ugly mix of semantically insignificant characters, the result will be made up entirely of printable characters. The characters in the range $00_{16}$ through $FF_{16}$ that are control or whitespace characters are mapped to the code points beginning at $100_{16}$. Luckily the entire block U+0100–U+17F0 is printable characters. The transition between the middle and bottom rows of Example 6 exemplifies these transformations.

## E  Unicode character handling SynCode evaluation prompt and grammar

As discussed in Section 6, to evaluate SynCode's ability to handle non-ASCII Unicode characters, we conducted an experiment using emoji generation task. We selected a subset of the TweetEval emoji dataset Barbieri et al. (2020), filtering for three common emoji classes. The task required the model to generate exactly one emoji character in response to a given tweet, adhering to a constrained grammar specification. This evaluation directly tested SynCode's handling of multi-byte UTF-8 sequences, which was a limitation in earlier versions.

**Listing 3** Grammar specification for emoji generation task using Unicode escape sequences.

```
// Lark grammar to validate single emoji output
start: emoji

// Define the 3 emojis from the TweetEval emoji dataset
emoji: "😍" | "😂" | "😉"
```

The prompt template instructed the model to analyze tweets and respond with exactly one emoji from the allowed set (Listing 4). We evaluated this task across 100 examples from the TweetEval test set on two versions of SynCode: v0.2.0 which used a character-level finite state machine (FSM), and v0.3.0, which implements a byte-level FSM with our recommended fix.

**Listing 4** Prompt template for emoji generation task (emoji symbols represented as placeholders)

```
You are evaluating tweets to assign the most appropriate emoji.
INSTRUCTIONS:
1. Read the tweet below carefully.
2. Select the SINGLE most appropriate emoji that captures the sentiment.
3. Respond with ONLY that emoji - no words or other characters.

The emoji must be one of the 3 valid options from this set:
😍 😂 😉

Tweet: "tweet_text"
Your response:
```

---

[4] A notable exception is OpenAI, who after GPT-2 have worked directly with byte sequences that are not guaranteed to be valid UTF-8.